\documentclass[preprint]{vgtc}               

\ieeedoi{10.1109/UncertaintyVisualization68947.2025.00010}

\title{Uncertainty-Aware PCA for Arbitrarily Distributed Data \\ Modeled by Gaussian Mixture Models} 

\author{%
    \authororcid{Daniel Klötzl}{0000-0002-4222-3320}\thanks{University of Stuttgart - firstname.lastname@visus.uni-stuttgart.de}
\and
    \authororcid{Ozan Tastekin}{0009-0007-8566-8745}\footnotemark[1]
\and
    \authororcid{David Hägele}{0000-0002-2679-6882}\footnotemark[1]
\and
    \authororcid{Marina Evers}{0000-0003-3904-5065}\thanks{University of Siegen - marina.evers@uni-siegen.de}
\and
    \authororcid{Daniel Weiskopf}{0000-0003-1174-1026}\footnotemark[1]
}

\abstract{
Multidimensional data is often associated with uncertainties that are not well-described by normal distributions. In this work, we describe how such distributions can be projected to a low-dimensional space using uncertainty-aware principal component analysis (UAPCA). We propose to model multidimensional distributions using Gaussian mixture models (GMMs) and derive the projection from a general formulation that allows projecting arbitrary probability density functions. The low-dimensional projections of the densities exhibit more details about the distributions and represent them more faithfully compared to UAPCA mappings. Further, we support including user-defined weights between the different distributions, which allows for varying the importance of the multidimensional distributions. We evaluate our approach by comparing the distributions in low-dimensional space obtained by our method and UAPCA to those obtained by sample-based projections.
}


\teaser{
  \centering
  \includegraphics[width=\linewidth]{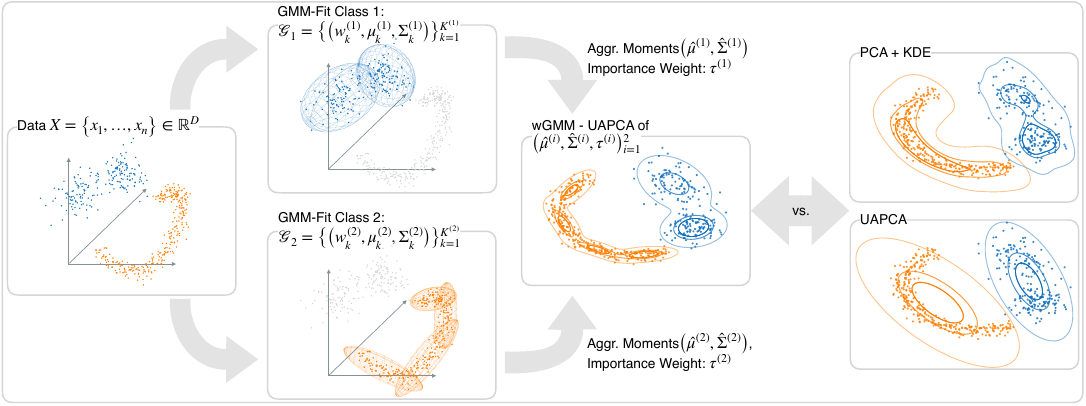}
  \caption{Overview of our uncertainty-aware PCA approach for an artificial dataset that consists of two classes of non-normally distributed samples. Each class is fitted with individual GMMs $\mathcal{G}_\text{\textit{i}}$ and aggregated for the projection via weighted UAPCA (with given weights $\tau^{\text{(\textit{i})}}$).
  Our approach is compared to PCA with KDE on the samples 
  and UAPCA.
  }
  \label{fig:teaser}
}

\graphicspath{{figs/}{figures/}{pictures/}{images/}{./}}

\usepackage{xcolor}
\usepackage[T1]{fontenc}
\usepackage[utf8]{inputenc}
\usepackage{amsmath, amssymb}
\usepackage[ruled,linesnumbered,vlined]{algorithm2e}
\usepackage{array, booktabs, makecell}
\usepackage{url}
\usepackage{cite}
\usepackage{xspace}

\newcommand{\abs}[1]{\left\lvert#1\right\rvert}
\newcommand{\diff}{\mathop{}\!\mathrm{d}}
\newcommand{\approach}{\ifmmode\mathrm{wGMM\mbox{-}UAPCA}\else\textnormal{wGMM-UAPCA}\fi\xspace}
\newcommand{\direct}{\ifmmode\mathrm{PCA\mbox{-}KDE}\else\textnormal{PCA-KDE}\fi\xspace}
\newcommand{\uapca}{\ifmmode\mathrm{UAPCA}\else\textnormal{UAPCA}\fi\xspace}
\newcommand{\kl}{\ifmmode\mathrm{KL}\else\textnormal{KL}\fi\xspace}
\newcommand{\wass}{\ifmmode\mathrm{SW}_2\else\textnormal{SW}\textsubscript{2}\fi\xspace}
\definecolor{owndarkblue}{rgb}{0.063, 0.435, 0.737}
\definecolor{ownred}{rgb}{0.894, 0.102, 0.110}

\usepackage{mathptmx}
\usepackage{csquotes}

\begin{document}
\firstsection{Introduction}

\maketitle
Principal Component Analysis (PCA) is a popular linear dimensionality reduction technique that has been used extensively in the literature to visualize high-dimensional datasets~\cite{liu2017hidimsurvey,xia2022drsurvey}.
For the application of PCA to uncertain data, various extensions have been proposed~\cite{denoeux2004fuzzy,Goertler2020UncertaintyAwarePCA,zabel2024vipurpca}.
Uncertainty-aware PCA (\uapca)~\cite{Goertler2020UncertaintyAwarePCA} assumes that the dataset is uncertainty-afflicted and consists of random vectors (as opposed to deterministic vectors) representing uncertainty in each observation.
The technique takes the variabilities (first and second moments) of the data items into account when computing the covariance matrix of the dataset from which the principal components are derived.
This makes \uapca broadly applicable, as it does not rely on the distribution, but only on its first and second moments, i.e., mean and covariance, to compute the overall covariance matrix.

However, the general applicability only holds for the computation of the principal components required to map the data to the low-dimensional space.
\uapca's subsequent projection for the visual mapping of the random vectors is only able to represent the projected first and second moments of the random vectors as Gaussians.
These are then depicted with contour plots of the corresponding probability density function (PDF), or by showing samples drawn from the Gaussians in a scatter plot.
This limitation can lead to misinterpretations, especially when the random vectors are not well approximated by a normal distribution.
For example, a multimodal distribution would be poorly represented by a Gaussian that only encodes the mean and covariance but cannot capture the various peaks at the modes, see the comparison of \uapca and \approach as denoted in \cref{fig:teaser} for the blue contour lines.
Properties like tailedness or asymmetries like skewness also cannot be observed when showing a Gaussian as a surrogate for the actual distribution.

In our work, we show how to overcome the limitation of \uapca's projection for visualizing the distribution.
Instead of projecting the distribution's first and second moments onto the principal components, we propose projecting the PDF and provide the mathematical formalism for doing so.
We apply the PDF projection to Gaussian mixture models (GMMs) due to their ability to model multimodal distributions and their versatility to approximate arbitrary distributions.
In addition, they allow a closed-form solution of the projected PDF, which is critical for efficiently implementing the proposed method without resorting to numerical integration.
We call our approach \approach, short for weighted Gaussian mixture model-based uncertainty-aware PCA. An overview of our approach is illustrated in \cref{fig:teaser}. For illustration purposes, we projected from 3D to 2D.

In our qualitative and quantitative evaluation, we compare our \uapca projections of GMMs with Gaussian surrogates and ground truth data.
The results show that our approach provides excellent approximations of the Monte Carlo estimates of the exact distributions and superior representation compared to \uapca's regular projection and visual mapping.
Furthermore, we propose incorporating a weighting mechanism of the random vectors into \uapca that emphasizes individual distributions.
This can be used to model sample imbalance, e.g., when samples from one distribution are more likely to occur than from another.
We show how this mechanism can be used in an interactive environment with an analysis scenario.

Our contributions can be summarized as:
\begin{itemize}
    \item A mathematical formulation for projecting arbitrary PDFs using a \uapca projection.
    \item An explicit formulation and evaluation for projecting and visualizing PDFs described by GMMs.
    \item A user-defined importance weighting for different distributions in \uapca.
\end{itemize}

\section{Related Work}
Including uncertainty in the visualization process imposes an important challenge in visualization research~\cite{weiskopf2022uncertainty, maack2023uncertainty, kamal2021recent, hagele2022uncertainty, brodlie2012review}. Skeels et al.~\cite{skeels2008revealing} provide an overview of various perspectives on uncertainty across different domains. In this work, we focus on uncertainty propagation in the projection process of dimensionality reduction for different input distributions. Thus, we aim to achieve a more accurate visual representation of the probability distribution in high-dimensional space to support an analysis on a higher level of detail.

There exists a wide range of dimensionality reduction techniques for deterministic data without uncertainty~\cite{nonato2018multidimensional, van2009dimensionality}. Uncertainty has been incorporated in dimensionality reductions, for example, by using probabilistic approaches~\cite{tipping1999mixtures, ding2011bayesian}. Uncertainty-aware multidimensional scaling (MDS)~\cite{hageleUncertaintyAwareMultidimensionalScaling2022} extends the MDS algorithm to uncertain points. Our approach builds on \uapca~\cite{Goertler2020UncertaintyAwarePCA}, which extends PCA to uncertain data. \uapca visualizes normal distributions in low-dimensional space by showing ellipses that indicate the first and second order moments of the distribution. VIPurPCA~\cite{zabel2024vipurpca} uses automatic differentiation to include the uncertainty in the PCA projection. However, these previous approaches focus on visualizing normally distributed data and do not allow for visualizing more complex distributions in low-dimensional spaces.

While normally distributed data can be easily represented by the mean and covariance matrix, various uncertainty models exist for alternative distributions. Recently, nonparametric models gained attention in the visualization context~\cite{pothkowNonparametricModelsUncertainty2013a, athawaleIsosurfaceVisualizationData2016, athawaleUncertaintyVisualizationCritical2024}. However, different parametric or combined models have also been used~\cite{hazarikaUncertaintyVisualizationUsing2017, eversUncertaintyawareSpectralVisualization2025}. In this work, we will derive a generic method for projecting probability distributions that only requires a representation of the multidimensional PDF. We derive the explicit formulation for projecting GMMs as a flexible and frequently used parametric model~\cite{bishop_pattern_2006}.

There are recent visualization approaches that target the visual representation of GMMs~\cite{giesen2024whole}. Gray~\cite{lawonn2022gray} includes methods for visualizing GMMs with a special focus on providing the ability to investigate the individual components. PrismBreak~\cite{zahoransky2025prismbreak} aims at using interaction to find good projection dimensions. However, these approaches focus on understanding a single distribution modeled by a GMM, whereas our work targets the projection and visualization of multiple GMMs, where each models an uncertain data point or one of multiple distributions of a dataset. We also focus on visualizing the entire PDF, rather than focusing on the individual \mbox{GMM components}.

Standard PCA and \uapca usually weigh all components equally. However, if the distributions represent different classes of significantly different sizes, an equal weighting provides misleading results. This problem has been observed in different contexts. For example, in computer graphics, a weighting based on the amount of surface data points is used to obtain reliable directions of the eigenvectors of a shape~\cite{obbtree}.
In the context of interactive visualization, domain expertise can be leveraged in user interaction.
iPCA~\cite{jeong2009ipca} explores the use of interaction for understanding PCA projections. 
However, they do not work on distributions that represent the underlying data. While several of the concepts can be applied to our approach of visualizing projections of different distributions, we focus on the interaction with class sizes.

\section{Mathematical Background} 
\label{sec:background}
We now provide the mathematical foundations for our approach by revisiting basic probability theory and multivariate Gaussian distributions, which serve as the foundation for modeling uncertainty. 
We then review principal component analysis (PCA) and its extension to uncertainty-aware PCA (\uapca)~\cite{Goertler2020UncertaintyAwarePCA}.
This serves as the starting point for our \uapca generalization to arbitrarily distributed data (i.e., not necessarily normally distributed).

\subsection{Probability Theory}
\label{ssec:probability-theory}
In the presence of measurement errors, noise, or incomplete data, uncertainty arises in many data analysis scenarios.
The specific uncertainty of a data point can be modeled via a random vector $X= \left(X_1, \ldots, X_n\right)^T$ characterized through $n$ random variables $X_i$ on the same space.
A random vector $X$ follows a multivariate normal distribution, denoted as $X\sim \mathcal{N}(\mu,\Sigma)$, if its PDF is given by 
$$f_X(x)
= \frac{1}{(2\pi)^{D/2}\,|\Sigma|^{1/2}}\,
\exp\left(-\frac{1}{2}(x-\mu)^T\Sigma^{-1}(x-\mu)\right)\,,$$
where $\mu\in \mathbb{R}^n$ denotes the mean (expected) vector and $\Sigma\in \mathbb{R}^{n\times n}$ the symmetric positive semidefinite covariance matrix, encoding the variability and correlations of the uncertainty.

In real-world data, the uncertainty is often estimated from a set of samples or observations $x_1,\ldots x_N \in \mathbb{R}^n$ and the multivariate Gaussian distribution is approximated by empirical mean and covariance.

While the multivariate normal distribution is widely used, it does not always reflect reality, due to the Gaussian assumption of the data. 
Underlying distributions may be skewed, heavy-tailed, or multimodal, due to mixture processes or nonlinearities in the data.
To address this limitation, a natural generalization is to approximate distributions with weighted combinations of Gaussian components in the form of Gaussian mixture models (GMMs). 
In this case, the PDF of a GMM $Y$ depends on a number of weights $w_k\ge0$, $\sum_{k=1}^K w_k = 1$ and corresponding multivariate normal distributions.
The PDF is given by the following convex combination of weighted multivariate normal distributions $\mathcal{N}(\mu_k, \Sigma_k)$,
\begin{equation} \label{eq:gmm_def}
    f_Y(x)=\sum_{k=1}^K w_k \mathcal{N}(x\vert \mu_k, \Sigma_k)\,.
\end{equation}
The combination of multiple Gaussians offers increased model flexibility, especially for cases where the underlying distributions are multimodal or possess significant asymmetries. 
However, this generalization also introduces additional complexity because the number of GMM components $K$ must be chosen or estimated, which introduces a hyperparameter that significantly affects the model quality.
As a result, an inappropriate number of components may introduce overfitting or underfitting of the model to the underlying distribution. 
Furthermore, the optimization of the weights is achieved with an expectation-maximization algorithm that is sensitive to initialization and often converges to local optima.
Despite these challenges, GMMs inherit favorable mathematical properties from multivariate normal distributions and allow a well-defined transformation of the PDF under a linear map obtained from \uapca.

\subsection{Principal Component Analysis}
\label{ssec:pca}
Dimensionality reduction techniques generally aim to reduce the dimensions of a dataset.
In this context, linear PCA projects a dataset $\mathcal{X} = \{x_1, \dots, x_N\} \subset \mathbb{R}^D$ to a lower-dimensional subspace $\mathbb{R}^d$ with $d < D$, while preserving the maximum amount of variance.
This is achieved by computing the set of orthonormal eigenvectors $\{u_1, \dots, u_D\}$ of the sample covariance matrix $$\Sigma := \frac{1}{N-1} \sum_{i=1}^N (x_i - \mu)(x_i - \mu)^T \,,
$$ with $\mu:=\frac{1}{N} \sum_{i=1}^N x_i$ the empirical mean of the data.
The principal components (or axes) are defined as the largest $d$ eigenvalues of $\Sigma$ and their corresponding eigenvectors $u_i$.
Projecting the high-dimensional points onto the low-dimensional span of eigenvectors results in points with reduced dimensionality
$
y_i := U^T (x_i - \mu)$, where $ U = [u_1, \dots, u_d] \in \mathbb{R}^{D \times d}$.

This method is well applicable to regular data consisting of fixed numeric values. 
However, when the dataset becomes subject to uncertainty, i.e., data points carry information about their variability, the computation of the covariance matrix that captures the variance of the whole dataset needs to be adapted to the more general case.

\subsection{Uncertainty-Aware Principal Component Analysis}
\label{ssec:uapca}

Görtler et al.~\cite{Goertler2020UncertaintyAwarePCA} propose a generalization by applying PCA not to $N$ individual data samples, but $N$ probability distributions. 
They modify the covariance calculation to assess the expected deviation for each distribution.
This computation incorporates both the average internal uncertainty of the distributions and the variability of their means.
As long as the moments of probability distributions are well defined, the method is applicable to arbitrary distributions.
However, for the subsequent projection and interpretation, the formulation assumes each distribution is a multivariate Gaussian.

For given random vectors $X_i\sim\mathcal{N}(\mu_i, \Sigma_i)$, let $\bar{\mu}:= \frac{1}{N} \sum_{i=1}^N\mu_i$ denote the mean of the distribution means, then the uncertainty-aware covariance matrix is defined using the average covariance $\bar{\Sigma}=\frac{1}{N} \sum_{i=1}^N \Sigma_i$, the dispersion of the means $\Sigma_{\mu}:= \frac{1}{N}\sum_{i=1}^N \mu_i \mu_i^T$, and the centering matrix $C = \bar{\mu}\bar{\mu}^T$.
The combination of the above expressions results in the uncertainty-aware covariance matrix
\begin{equation}\label{eq:uapca_p}
\Sigma_{\mathrm{UA}} = \Sigma_{\mu} + \bar{\Sigma} - C\,.
\end{equation}
As in PCA, the eigenvalues and eigenvectors of $\Sigma_{\mathrm{UA}}$ are determined to construct the projection.
While the original \uapca approach projects the first and second moments of each distribution, it assumes that each projected distribution remains Gaussian.
To overcome this limitation, we extend \uapca by formulating the direct projection of PDF, especially for GMMs, in the following section.

\section{Generalization to Arbitrary Distributions}
This section introduces a general framework to propagate PDFs through the PCA projection. 
We evaluate the proposed ansatz by computing the projection for the special case of GMMs and present a concise visualization for the projected distributions.

\subsection{Projection of PDF}
\paragraph{Ansatz:} Given a random vector $X\in \mathbb{R}^D$ with PDF $f_X(x)$. Let $U := \begin{bmatrix}P & R\end{bmatrix}  \in \mathbb{R}^{D\times D}$ be an orthogonal matrix concatenating a projection matrix $P\in \mathbb{R}^{D\times d}$ and its orthogonal complement $R\in \mathbb{R}^{D\times D-d}$.
Applying $U$ to a point $x\in \mathbb{R}^D$ decomposes into the projection component $y\in \mathbb{R}^d$ and residual $z\in \mathbb{R}^{D-d}$ via
$$x=Py + Rz \quad \text{for } y=P^T x \text{ and } z=R^T x \,. $$
Marginalization of the orthogonal complement leads to the projected PDF of the random vector $Y:= P^T X$: 
\begin{equation}\label{eq:ansatz}
    f_Y(y) = \int_{\mathbb{R}^{D-d}}f_X(Py + Rz) \diff z\,.
\end{equation}
In general, this equation allows the projection of arbitrary PDFs.
In this paper, we present the usefulness of our ansatz by computing the solutions for PDFs represented as GMMs.

\subsection{UAPCA Projection for Arbitrarily Distributed Data -- Modeled via GMMs}
\label{ssec:nnuapca_gmm}
In the following, we apply the above approach to arbitrarily distributed data.
Let us assume that this data is modeled via GMMs.

\paragraph{Computation of Projected GMM PDF}
To apply the approach to GMMs, we substitute the PDF of a GMM given in \cref{eq:gmm_def} into \cref{eq:ansatz} and swap the final sum and integral:
\begin{align*}
f_Y(y)
&= \int_{\mathbb{R}^{D-d}}\sum_{k=1}^K w_k\,\mathcal{N}\Bigl(Py+Rz\mid \mu_k,\Sigma_k\Bigr)\,\diff z\\
&= \sum_{k=1}^K w_k\int_{\mathbb{R}^{D-d}}\mathcal{N}\Bigl(Py+Rz\mid \mu_k,\Sigma_k\Bigr)\,\diff z \,.
\end{align*}
By the result of the marginalization of individual multivariate normal distributions, discussed in detail in the \hyperref[appendix]{Appendix}, \cref{eq:exp_2}, each integral in the sum becomes $\mathcal{N}(y\mid P^T\mu_k,\,P^T\Sigma_k P)$. 
This yields
\begin{equation}
\label{eq:gmm_result}
f_Y(y) = \sum_{k=1}^K w_k\,\mathcal{N}\Bigl(y\mid P^T\mu_k,\,P^T\Sigma_k P\Bigr) \,.
\end{equation}
As a result, the PDF of the projected GMM is (again) characterized only by the specific weights $w_k$ and a set of projected mean vectors ($P^T\mu_k$) and covariance matrices ($P^T\Sigma_kP$).

\paragraph{Aggregated Mean and Covariance of GMMs}
In \uapca, a single mean vector and a covariance matrix are necessary as input. 
In our case, as discussed above, we have multiple mean vectors and covariance matrices for each of the GMM components.

To compute the uncertainty-aware projection matrix, the formulation requires a single aggregated mean vector $\hat{\mu}$ and an aggregated covariance matrix $\hat{\Sigma}$ for each of the uncertainty distributions.
In the case of a GMM with $K$ components, characterized by weights $w_k\in\mathbb{R}\ge0$, means $\mu_k\in \mathbb{R}^D$, and covariances $\Sigma_k \in \mathbb{R}^{D\times D}$, the weighted average of the component means is computed via
$\hat{\mu} = \sum_{k=1}^K w_k\mu_k$.

The computation of the covariance matrix is not restricted to summing the weighted covariances but takes into account the between-component variability (covariance between means of components) as well.
This results in the following formulation for the overall covariance matrix $\hat{\Sigma}$:
$$
\hat{\Sigma} = \sum_{k=1}^{K} w_k \Sigma_k
+ \;
\sum_{k=1}^{K} w_k (\mu_k - \hat{\mu})(\mu_k - \hat{\mu})^T \,.
$$
This calculation of the overall mean and covariance follows Bishop~\cite[Eqs.~(9.49) and (9.50)]{bishop_pattern_2006}.
With the aggregated quantities, the uncertainty-aware covariance matrix $\Sigma_{\mathrm{UA}}$ is computed, and its leading eigenvectors form the projection matrix $P$, see~\Cref{ssec:uapca}.

\subsection{Weighted Uncertainty-Aware Projection Matrix}
\label{ssec:weighted_uapca_p}
The previous section discussed the computation of aggregated means and covariances of given GMMs that can be used to generate the uncertainty-aware projection matrix $P$.

A natural extension of this framework is the incorporation of importance weights $\tau$ into the computation of the projection matrix.
This is particularly relevant in scenarios where the distributions originate from sample-based modeling.
In this case, class imbalance, i.e., unequal numbers of samples per class, may be present in the original data but gets neglected once the classes are modeled by distributions and treated equally in the \uapca projection process.

In the special case discussed above, where GMMs are fitted to individual data classes, a canonical choice could be to assign weights proportional to the number of samples.
This would lead to distributions derived from larger populations having a greater impact on the resulting projection.
In the following, we describe the integration of weights into the \uapca framework in the context of GMMs following \cref{alg:weighted_uapca_projection}. 

We reconsider \cref{eq:uapca_p} described in \Cref{ssec:uapca} and reformulate the components by the above introduced importance weights.
First, we combine the GMMs of each class with the aggregated means and covariances introduced in the previous section.
As formalized in \cref{alg:aggregate_means} -- \cref{alg:aggregate_covs}, this process reduces each $\mathcal{G}^{(i)}$ to
$$
\left\{ w_k^{(i)},\mu^{(i)}_k, \Sigma_k^{(i)} \right\}_{k=1}^{K^{i}} \longrightarrow  \left(\hat{\mu}^{(i)}, \hat{\Sigma}^{(i)}\right)\,,
$$
with corresponding importance weights $\tau^{(i)}$.
Using these weights, we define the components of the uncertainty-aware formulation as:
\begin{align*}
	\bar{\mu}&:=\sum_{i=1}^L \tau^{(i)}\cdot \hat{\mu}^{(i)}\,, &\qquad C:&=\bar{\mu}\bar{\mu}^T\,,\\
	\bar{\Sigma}&:= \sum_{i=1}^L \tau^{(i)} \cdot \hat{\Sigma}^{(i)}\,, &\qquad \Sigma_{\hat{\mu}}:&= \sum_{i=1}^L \tau^{(i)} \cdot \hat{\mu}^{(i)} (\hat{\mu}^{i})^T\,.
\end{align*}
This enables the analogous computation of the weighted uncertainty-aware covariance matrix $\Sigma_{\mathrm{UA}} = \Sigma_{\hat{\mu}} + \hat{\Sigma} - C$,
which generalizes the unweighted case and coincides with the unweighted version when all weights $\tau^{(i)}$ are set to $1/L$.
Finally, the uncertainty-aware covariance matrix is decomposed using eigendecomposition, and the projection matrix $P\in \mathbb{R}^{D\times d}$ results from the corresponding eigenvectors of the largest $d$ eigenvalues.

Since we will evaluate our approach for real-world sample-based datasets, we use the weighted uncertainty-aware projection throughout \Cref{ssec:eval_real_world}.
Beyond sample-based weighting, the integration of weights $\tau$ into the uncertainty-aware projection matrix allows a user-driven interactive system.
Users can manually specify or adapt weights to emphasize or de-emphasize certain distributions or classes.
This will be further demonstrated in \Cref{ssec:eval_weights}.

\begin{algorithm}[!t]
	\caption{Computation of the Weighted Uncertainty-Aware Projection Matrix}
	\label{alg:weighted_uapca_projection}
	
	\SetKwInOut{Input}{Input}
	\SetKwInOut{Output}{Output}
	\Input{
		$\{\mathcal{G}^{(i)}\}_{i=1}^L$ -- GMMs for $L$ classes, where $\mathcal{G}^{(i)} = \{(w_k^{(i)}, \mu_k^{(i)}, \Sigma_k^{(i)})\}_{k=1}^{K^{(i)}}$, \\
		$\{\tau^{(i)}\}_{i=1}^L$ -- importance weights for each class
	}
	\Output{
		Projection matrix $P \in \mathbb{R}^{D \times d}$
	}
	
	\BlankLine
	\textbf{Step 1: Aggregate GMM Means and Covariances}\\
	\For{$i= 1$ \KwTo $L$}{\label{alg:aggregate_means}
		Compute aggregated mean: $\hat{\mu}^{(i)} = \sum_{k=1}^{K^{(i)}} w_k^{(i)} \mu_k^{(i)}$ \\
		Compute aggregated covariance: \\
        $\hat{\Sigma}^{(i)} =
        \sum_{k=1}^{K^{(i)}} w_k^{(i)}
        \left[ \Sigma_k^{(i)} + (\mu_k^{(i)} - \hat{\mu}^{(i)})
        (\mu_k^{(i)} - \hat{\mu}^{(i)})^T \right]$\label{alg:aggregate_covs}
	}
	
	\BlankLine
	\textbf{Step 2: Compute Weighted Aggregated Quantities} \\
	Compute weighted mean: $\bar{\mu} = \sum_{i=1}^L \tau^{(i)} \cdot \hat{\mu}^{(i)}$\label{alg:weighted_mean} \\
	Compute centering matrix: $C = \bar{\mu} \bar{\mu}^T$ \\
	Compute weighted covariance matrix: $\bar{\Sigma} = \sum_{i=1}^L \tau^{(i)} \cdot \hat{\Sigma}^{(i)}$ \\
	Compute weighted mean outer product: $\Sigma_{\hat{\mu}} = \sum_{i=1}^L \tau^{(i)} \cdot \hat{\mu}^{(i)} (\hat{\mu}^{(i)})^T$\label{alg:weighted_mean_outer}
	
	\BlankLine
	\textbf{Step 3: Compute Uncertainty-Aware Covariance Matrix} \\
	$\Sigma_{\mathrm{UA}} = \Sigma_{\hat{\mu}} + \bar{\Sigma} - C$
	
	\BlankLine
	\textbf{Step 4: Compute Projection Matrix} \\
	Compute eigendecomposition of $\Sigma_{\mathrm{UA}}$ \\
	Set $P \in \mathbb{R}^{D \times d}$ the leading $d$ eigenvectors of $\Sigma_{\mathrm{UA}}$
\end{algorithm}

\begin{table*}\setlength{\tabcolsep}{5pt}
\caption{Quantitative evaluation of 17 real-world labeled datasets. Each of the dataset classes (defined by the distinct labels) was modeled by GMMs with the specified number of GMM components. 
The three approaches, 
wGMM-UAPCA, UAPCA, and PCA-KDE,
are quantitatively evaluated using the Kullback-Leibler divergence (KL)
and sliced Wasserstein distance (SW$_\text{2}$).
}
\label{tab:espadoto_quant_eval}
\begin{tabular}{lrrcccccc}
\toprule
Dataset & \makecell{ Size\\ } & \makecell{Dim. \\  } & \makecell{Labels \\  } & \makecell{GMM \\ Components} & \makecell{ \kl{}(\direct, \\ \approach) } & \makecell{ \kl{}(\direct, \\ \uapca) } & \makecell{ \wass{}(\direct, \\ \approach) } & \makecell{ \wass{}(\direct, \\ \uapca) }\\
\midrule
bank & 2058 & 63 & 2 & \makecell{16, 8} & \textbf{4.0804e-04} & 8.1839e-04 & \textbf{1.2428e-02} & 1.9343e-02 \\
cifar10 & 3249 & 1024 & 10 & \makecell{1, 14, 1, 1, 1 \\ 1, 1, 13, 12 14} & 3.6982e-03 & \textbf{3.1093e-03} & \textbf{9.6107e-03} & 1.0517e-02 \\
cnae9 & 1079 & 856 & 9 & \makecell{2, 1, 1, 1, 2 \\ 2, 1, 1, 1} & \textbf{4.3181e-04} & 4.4680e-04 & 8.9226e-03 & \textbf{8.8375e-03} \\
coil20 & 1439 & 400 & 20 & \makecell{3, 3, 3, 3, 3 \\ 3, 3, 2, 3, 2 \\ 3, 2, 3, 3, 2 \\ 1, 2, 3, 3, 3} & 3.9615e-03 & \textbf{2.6747e-03} & \textbf{8.6938e-03} & 1.2066e-02 \\
epileptic & 5749 & 178 & 5 & \makecell{2, 3, 2, 1, 1} & \textbf{3.7940e-05} & 6.0040e-05 & \textbf{5.2632e-03} & 6.1744e-03 \\
fashion\_mnist & 2999 & 784 & 10 & \makecell{1, 2, 12, 10, 11 \\ 2, 1, 2, 10, 2} & \textbf{8.4062e-03} & 9.9694e-03 & \textbf{8.8474e-03} & 1.2469e-02 \\
fmd & 996 & 1536 & 10 & \makecell{4, 4, 4, 4, 4 \\ 4, 5, 4, 3, 5} & 1.5755e-03 & \textbf{9.4136e-04} & 1.4576e-02 & \textbf{1.2633e-02} \\
har & 734 & 561 & 6 & \makecell{5, 5, 4, 4, 4, 5 \\} & 3.0825e-03 & \textbf{2.2314e-03} & \textbf{8.5085e-03} & 1.1451e-02 \\
hatespeech & 3220 & 100 & 3 & \makecell{4, 9, 5} & \textbf{1.8865e-04} & 3.1003e-04 & \textbf{1.3642e-02} & 2.2184e-02 \\
hiva & 3075 & 1617 & 2 & \makecell{6, 5} & \textbf{1.2059e-03} & 3.2603e-03 & \textbf{7.5815e-03} & 1.5401e-02 \\
imdb & 3249 & 700 & 2 & \makecell{1, 1} & 1.7140e-05 & 1.7140e-05 & \textbf{1.0383e-02} & 1.0778e-02 \\
secom & 1566 & 590 & 2 & \makecell{2, 6} & \textbf{6.7170e-04} & 6.9461e-04 & \textbf{2.4978e-02} & 4.2688e-02 \\
seismic & 645 & 24 & 2 & \makecell{8, 1} & 9.2591e-03 & \textbf{1.1034e-03} & \textbf{1.8918e-02} & 2.2457e-02 \\
sentiment & 2747 & 200 & 2 & \makecell{13, 16} & \textbf{7.2360e-05} & 2.5743e-04 & \textbf{1.3689e-02} & 4.0064e-02 \\
sms & 834 & 500 & 2 & \makecell{2, 1} & \textbf{3.4828e-04} & 3.5066e-04 & \textbf{3.1853e-02} & 3.3163e-02 \\
spambase & 4600 & 57 & 2 & \makecell{15, 7} & \textbf{1.0266e-04} & 2.0145e-04 & \textbf{1.6376e-02} & 2.4708e-02 \\
svhn & 731 & 1024 & 10 & \makecell{6, 3, 3, 3, 3 \\ 3, 3, 2, 2, 2} & 1.1899e-02 & \textbf{5.7717e-03} & \textbf{1.2997e-02} & 1.4286e-02 \\
\bottomrule
\end{tabular}
\end{table*}

\subsection{PDF Contour Line Visualization}
\label{ssec:pdf_visualization}
The visual representation of projected PDFs in the two-dimensional space is necessary to interpret and compare the different densities per label.
In prior work, (projected) distributions are typically visualized using probability samples, isobands, or isolines at specific probability levels. 
While conveying the distribution extent, probability samples tend to introduce visual clutter, and isobands may cause overdraw, particularly for overlapping classes.
We therefore employ isolines at selected probability levels (e.g., 25th, 50th, and 95th percentiles), which reduce overdraw and preserve the distribution shape.
This representation supports visual comparison of class distributions and enables tasks such as clustering and outlier detection.

In the original \uapca method~\cite{Goertler2020UncertaintyAwarePCA}, contour lines representing specific levels of the cumulative density function are defined by ellipses fully determined by the distribution's mean and covariance matrix. 
In comparison, this direct approach does not hold true for the projected GMMs anymore.
We describe a three-step numerical approach to identify contour lines for two-dimensional GMMs.

\begin{figure}[!b]
    \centering
    \includegraphics[width=\linewidth]{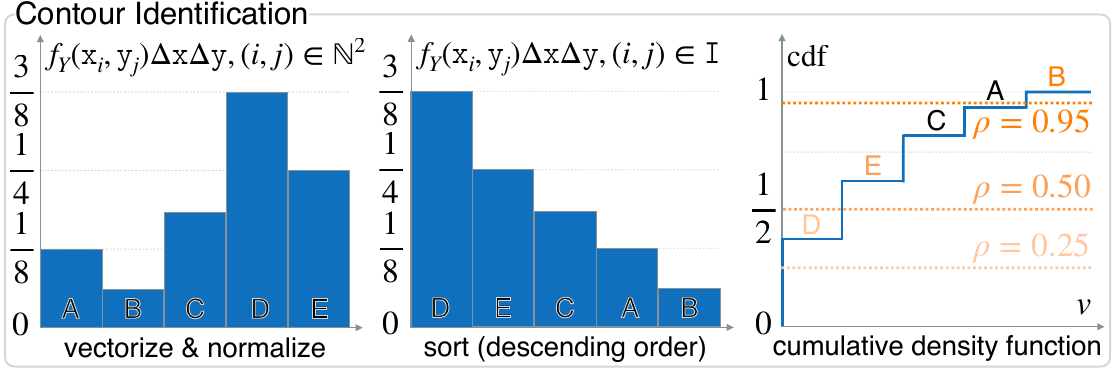}
    \caption{Contour identification follows a three-step process: vectorize the flattened PDF values (left), sort them in descending order (middle), and compute the cumulative density function (cdf) by summation of the sorted values (right). For illustration, discrete PDF values are labeled alphabetically (A, B, ...). The dotted colored lines indicate contour levels at $\rho \in \{25\%, 50\%, 95\%\}$, highlighting grid cells whose cdf exceeds the corresponding thresholds.}
    \label{fig:threshold_example}
\end{figure}

Let us assume that a discrete PDF $f_Y(\mathtt{x}_i, \mathtt{y}_j), \, (i,j)\in \mathbb{N}^2$ on a uniform two-dimensional grid with spacing $\Delta \mathtt{x}$ and $\Delta \mathtt{y}$ is given.
For reference, we provide an illustrative example of the contour line extraction approach in \cref{fig:threshold_example}.
This shows the one-dimensional process, e.g., after vectorization of the two-dimensional grid.
To comply with the PDF property that the total integral (sum) over the whole domain is one, the first step is the normalization of the PDF values, shown in the left part of~\cref{fig:threshold_example} for a bar width of $\Delta \mathtt{x} \cdot\Delta \mathtt{y} = 1$.
In the second step, the density values $f_Y(\mathtt{x}_i,\mathtt{y}_j)$ are sorted in descending order (middle) and the permutation of indices is denoted via $\mathtt{I}$. 
To identify the contour lines, the cumulative density function (cdf) of the sorted density values $f_Y(\mathtt{x}_i,\mathtt{y}_j)$, $(i,j)\in \mathtt{I}$ is computed by accumulating the top density values
$$
\operatorname{cdf}(v) = \sum_{(i,j)\in  \mathtt{I}_v} f_Y(\mathtt{x}_i, \mathtt{y}_j) \; \Delta \mathtt{x}\; \Delta \mathtt{y} \,,
$$
where $\mathtt{I}_v$ are the first $v$ indices in the sorted order.
As a result, $\operatorname{cdf}(v) > 0.5$ indicates that the first $v$ density values in the sorted list collectively cover more than 50\% of the total density mass.

The above-described scheme enables the identification and visualization of specific quantiles of the PDF, such as the proposed $\rho \in \{25\%, 50\%, 95\%\}$, by determining all grid cells where the cdf is greater than or equal to $\rho$.
In the projection visualizations of the GMMs, the three contours cover the tail regions, the spread, and the core(s), respectively.
The different levels are differentiated by using opaque colored lines (colored linearly by the respective $\rho$-values).

\subsection{Application Scenarios}
\label{ssec:application_scenarios}

In real-world applications of dimensionality reduction, data uncertainty often arises in one of the following two forms.
Either it is inherently provided by a finite ensemble of possible realizations of an underlying distribution or explicitly defined via PDFs, e.g., uniform distribution, multimodal, or otherwise structured.
The first case is evaluated in \Cref{tab:espadoto_quant_eval}, where we apply our method to a diverse collection of datasets.
The second case covers scenarios where uncertainty is given via analytical PDFs.

\paragraph{Uncertainty-Aware PDFs from Realizations}

When uncertain data is provided as ensembles, e.g., multiple samples per class or simulation outputs, a common approach is to fit GMMs to the data of each class or label.
Following the formulation in \Cref{ssec:weighted_uapca_p}, we compute the aggregated mean and covariance for each GMM and apply the (weighted) \uapca method to obtain the projection matrix~$P$.
This enables dimensionality reduction that preserves the derived uncertainty characteristics.
Hence, the generalization to GMM-based modeling offers an interpretable alternative, compared to the direct PCA projection of an infeasible number of initial samples.

\paragraph{Analytic Uncertainty-Aware PDFs}
Our approach also accommodates analytically defined uncertainty distributions, where the PDFs are known in closed form. 
In engineering or physics applications, for example, distributions over parameter spaces may be explicitly defined based on theoretical models or prior knowledge.
As discussed in the \hyperref[appendix]{Appendix} and \Cref{ssec:nnuapca_gmm}, this integration reduces to closed-form expressions when $f_X(x)$ is a multivariate normal distribution or a GMM.
Therefore, in these cases, no sampling or modeling step is required.
For arbitrarily distributed PDFs, however, we propose the following procedure to match the underlying distribution as closely as possible and compute a suitable PCA projection.
Given the analytically expressed PDF, a number of samples from the distributions are picked.
We can use these samples to fit a new GMM in the high-dimensional space and represent the non-normal distribution via this GMM. 
This approach is evaluated in \Cref{ssec:eval_students} for a given analytic dataset.

\subsection{Fitting Gaussian Mixture Models to Sampled Data}
\label{ssec:fitting_gmms_samples}
In both of the previously mentioned application scenarios, the fitting of Gaussian mixture models to samples is a key preprocessing step, before we can apply \approach. 
We employ the Bayesian Information Criterion (BIC) to identify the optimal number of components within a given range.
The BIC balances model fit and complexity by penalizing models with more parameters, thus helping prevent overfitting.
Since direct BIC evaluation in the original high-dimensional space is unstable and, in this case, dominated by the penalty terms that scale with the number of estimated parameters, we use PCA on the given dataset samples to reduce the dimensionality to lower dimensions.
This was motivated by Lee and Verleysen~\cite{lee2007nonlinear} as described in the context of nonlinear dimensionality reduction. 
They stated that using PCA to reduce a large number of dimensions to 50--200 dimensions improves the subsequent (nonlinear) dimensionality reduction methods. 
The GMM is then fitted in this lower-dimensional space across a range of GMM components, and the configuration with the lowest BIC value is selected.
Finally, the chosen number of components is used to fit the GMM in the original high-dimensional feature space.

\begin{figure*}[!t]
    \centering
    \includegraphics[width=\linewidth]{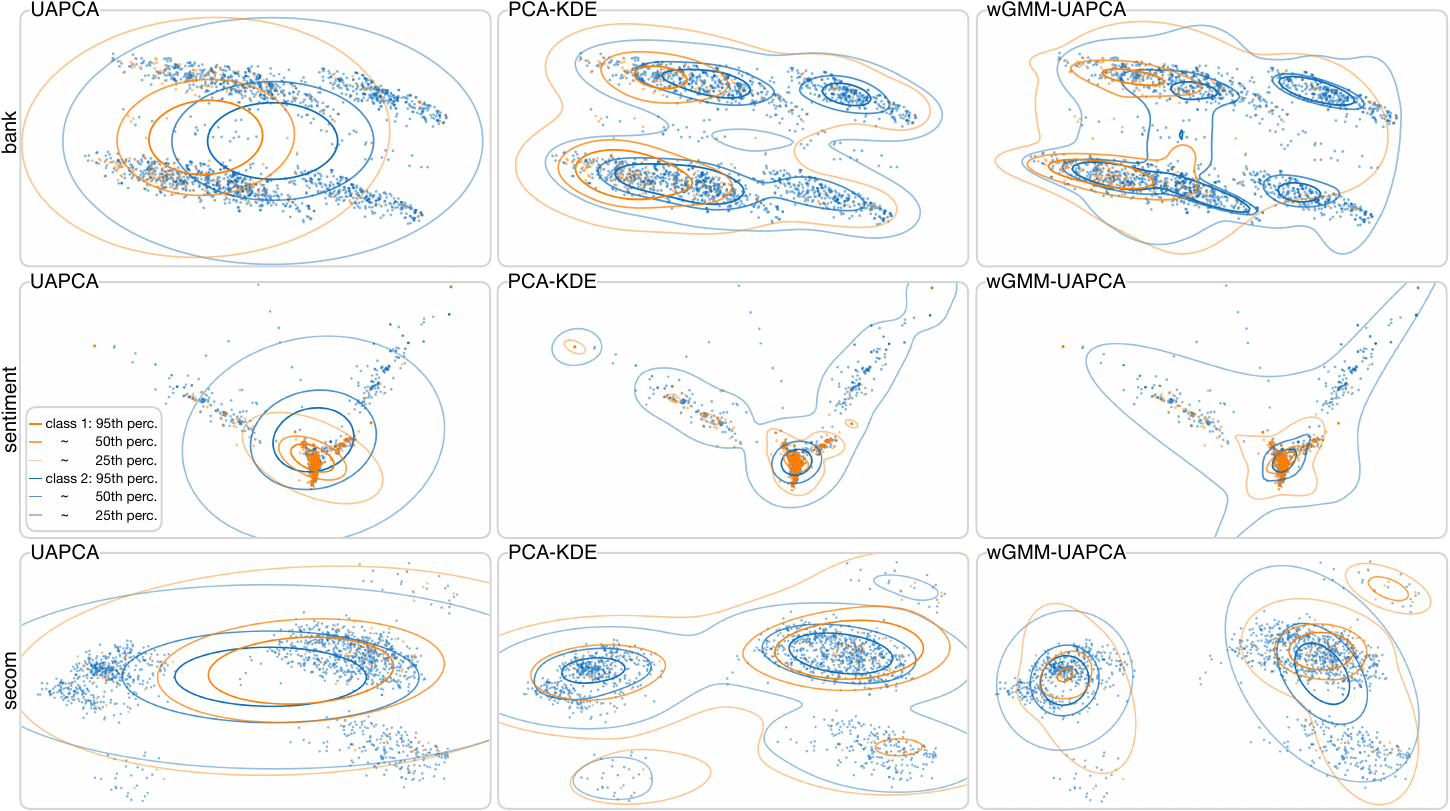}
    \caption{Visualizations of the projections for different datasets and projection methods. Each dataset has two classes, and the projection of the original sample points is shown as a reference, together with contour lines representing the isolines of the PDF. For demonstration and comparison, we use the wGMM-UAPCA projection matrix to consistently project each dataset, i.e., the samples, multivariate normal distributions, and GMMs.}
    \label{fig:espadoto_eval_qual}
\end{figure*}

\section{Evaluation}
Our approach is evaluated in several different respects.
First, the approach is applied to real-world datasets~\cite{Espadoto2021} and evaluated quantitatively and qualitatively.
An overview of the datasets and the quantitative evaluation is given in \Cref{tab:espadoto_quant_eval}.
In this setting, labeled data samples are modeled using GMMs, and GMM-based PCA is applied.
Secondly, the approach is applied to a dataset given as explicit distributions that we approximate by GMMs.
Third, the effectiveness of the user-specific weighted \approach projection is demonstrated for two datasets.

All computations were performed in a Jupyter Notebook environment on a MacBook Pro equipped with an Apple M4 Pro chip and 24\,GB of RAM. 
The code and evaluation specifics are publicly available.\footnote{\url{https://github.com/VisDan93/wGMM-UAPCA}}
The core method will be integrated in the uncertainty-aware data analysis Python package \textsc{UADAPy}~\cite{UADAPy}.

\begin{figure*}
    \centering
    \includegraphics[width=\linewidth]{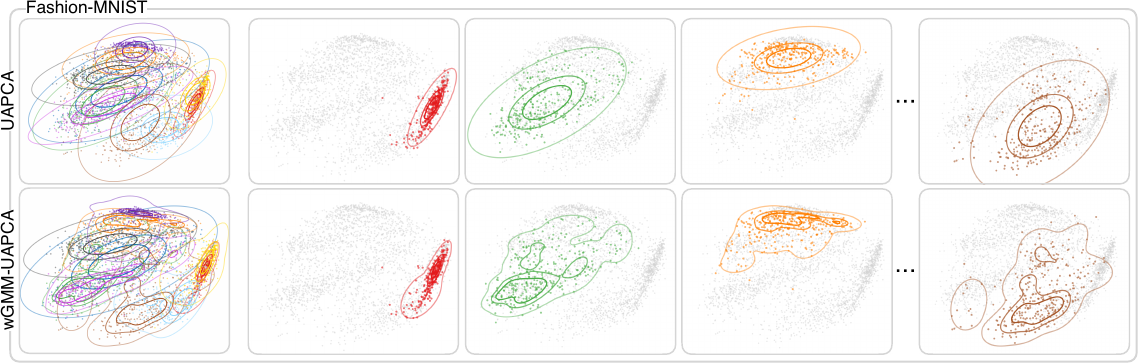}
    \caption{The projection of the Fashion-MNIST dataset with 10 classes. The first column shows all classes together for
     UAPCA 
     (top) and wGMM-UAPCA 
     (bottom), while the following columns show only individual classes to avoid overplotting.}
    \label{fig:fashion_mnist_eval}
\end{figure*}

\subsection{Evaluation of Real-World Datasets}
\label{ssec:eval_real_world}

For the evaluation of the introduced \approach, we use the set of real-world labeled datasets described by Espadoto et al.~\cite{Espadoto2021}. 
It consists of 18 real-world datasets with varying dimensionality, samples, and number of labels. 
As outlined in \Cref{ssec:application_scenarios}, each class in each dataset is modeled by a GMM.
In the data from their project website,\footnote{\url{https://mespadoto.github.io/proj-quant-eval/post/datasets/}(accessed 2025-07-17).} the dataset orl is omitted in our evaluation since classes consist of only single samples.

To ensure reproducibility across datasets, we adopt an automated and consistent procedure for determining the number of GMM components per class.
Instead of manual or dataset-specific tuning, we use the procedure discussed in \Cref{ssec:fitting_gmms_samples}.
For each class, we fit GMMs with 1--30 components and select the configuration with the lowest BIC.
If the dimensionality is high ($D>50)$, we first reduce the data via PCA to $50$ dimensions and perform the model selection in this reduced space.
The selected number of components is then used to fit the final GMM in the original feature space.
The respective number of components is reported in \Cref{tab:espadoto_quant_eval}.

To assess the quality of our projection approach, we compare three strategies for visualizing uncertain data:

\begin{itemize}
	\item \textbf{\uapca:} Each labeled sample set is modeled as a multivariate normal distribution using empirical mean and covariance. The resulting Gaussian distributions are then projected using (weighted) \uapca, yielding a projected Gaussian PDF.
	
	\item \textbf{\approach:} A GMM is fitted to the labeled sample set using the determined number of components. The fitted GMM is then projected using the weighted projection described in \Cref{ssec:nnuapca_gmm}, resulting in a projected mixture distribution.
	
	\item \textbf{\direct:} The samples are projected directly using PCA. A kernel density estimate (KDE) is then computed in the low-dimensional space to approximate the resulting distribution.
\end{itemize}

We choose \direct as reference because it minimizes assumptions after the projection. 
Another option would be to fit a GMM to the projected samples, which would introduce additional parameters (number of GMM components) and could favor models that rely on GMM assumptions.

For each dataset, we conduct a quantitative evaluation by comparing the projected PDFs produced by \uapca and \approach against the KDE-based reference.
To cover approximation quality, we use both the Kullback-Leibler (\kl) divergence and the sliced \mbox{2-Wasserstein} distance (\wass).
Results are summarized in \Cref{tab:espadoto_quant_eval} and discussed below, followed by a qualitative analysis of the projected PDFs for selected datasets in~\cref{fig:espadoto_eval_qual}.

\paragraph{Quantitative Evaluation using KL Divergence and Sliced Wasserstein Distance}
To get started, we briefly discuss the two metrics used for measuring distances between probability distributions (\kl and \wass). The \kl~divergence is defined for two probability distributions $p$ and $q$ via $$\kl(p\Vert q) = \int p(x) \log\left( \frac{p(x)}{q(x)}\right)\diff x\,.$$
This formula implies an asymmetric behavior and is interpreted as measuring the information loss when $q$ is used to approximate $p$. 
In our evaluation, we use the KDE result (\direct) as the reference distribution $p$ and compare against the approximated PDFs from $\uapca$ and $\approach$.
While \kl captures pointwise differences in probability mass and is sensitive to small mismatches in high-probability regions, it may penalize slight shifts of the distributions. This leads to less robust measurements when distributions have disjoint support.

To complement this behavior, we also compute the sliced Wasserstein distance $\wass$, an efficient variant of the Wasserstein distance. 
This distance measures intuitively how much mass needs to be moved to transform one distribution into another.
Compared to the \kl distance, this measure has metric properties: it is symmetric and respects the triangle inequality.
The used implementation from the Python Optimal Transport (POT) library~\cite{flamary2021pot} was performed with a fixed set of 5000 samples per distribution.

In our setting, we computed \kl~divergence and \wass~distance for each label of the datasets. 
This required the combination of the metrics that was performed using weighted averages over the classes.
Thereby, distortion due to class imbalance is avoided and the evaluation is aligned with the weighting applied in \approach.

Overall, the \kl divergence shows better results (lower \kl values marked in bold) for the comparison between \direct and our \approach for 10 out of the 17 datasets, suggesting an improved match of the projected distributions. 
The \wass distance (lower \wass values marked in bold) confirms and even emphasizes this insight. 
However, for the fmd dataset, both the \kl and \wass are worse compared to \uapca. 
This could hint at GMM overfitting or poorly separable Gaussians for the classes.
An interesting case is the seismic dataset, where \uapca achieves a much lower \kl divergence compared to \approach, whereas the \wass is clearly lower for \approach.
This ambiguity could hint at either both approaches being inaccurate, or our assumed ground-truth \direct missing the true underlying distribution.

Lastly, in datasets like bank or sentiment, \approach significantly outperforms \uapca in both metrics, accurately capturing class-specific multimodality and asymmetries that \uapca cannot represent.
We will look into these dataset projections (illustrated in \cref{fig:espadoto_eval_qual}) in more detail in the following paragraph.

As expected, in imdb, where each class is modeled with a single Gaussian, both \uapca and \approach yield nearly identical results (the small difference in \wass is related to the sample-based computation of the Wasserstein distance).

To sum up, these findings confirm that while being dependent on the GMM model, our method achieves more faithful projection representations of complex classes affected by uncertainty in a variety of real-world datasets.
Moreover, the core computations for \approach, i.e., determining the projection matrix and applying it to the GMMs to obtain the projected PDFs, require between 4\,ms and 200\,ms depending on the dataset.
This enables responsive, interactive analysis alongside improved accuracy. 
For comparison, the computations for \uapca were in the range of 1\,ms to 150\,ms for the same datasets.

\begin{figure*}[!t]
    \centering
    \includegraphics[width=\linewidth]{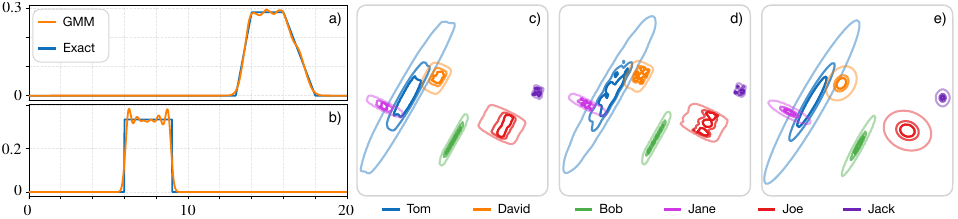}
    \caption{The student dataset's trapezoidal (a, for student David and subject M2) and uniform (b, for student Jack and subject P2) distributions can be better approximated by GMMs than normal distributions. Thus, the KDE of the projected points (c) can be better approximated by the GMM projection (d) than by UAPCA
    (e).}
    \label{fig:student_eval}
\end{figure*}

\paragraph{Qualitative Evaluation of Projections of Real-World Datasets}
In the following, we visually compare the projections to obtain a better understanding of the strengths and weaknesses of the projection approaches. For this comparison, we use the \approach projection for both the samples and the distributions. Note that our projection coincides with the direct sample-based PCA projection. \cref{fig:espadoto_eval_qual} shows the results for three different datasets where the individual classes deviate from normal distributions.

At first, we investigate the bank dataset, which consists of two overlapping multimodal distributions separable in the PCA projection. As \uapca only projects the mean and variance, the multimodality cannot be captured. In contrast, our approach effectively captures the multimodality. When comparing to the samples of the original data or their KDE, we observe that the internal structure is similar.
Complementary observations can be made for the sentiment dataset. As for the bank dataset, the quantitative evaluation yielded significantly better results for \approach compared to \uapca. This can also be confirmed visually (see \cref{fig:espadoto_eval_qual}). Visualizing mean and covariance in \uapca does not capture the two separate branches in the low-dimensional distribution of the blue class. Our approach captures these features well, even though we observe that some contour lines of the GMM lead to an increased density in regions with few or no samples (see bottom-left spike).

The third example shows two overlapping distributions for the secom dataset. When observing the results for \approach in \cref{fig:espadoto_eval_qual}, we notice that the GMM components do not accurately describe the distributions. For example, the sample points in the lower left are not present in our GMM. However, this is most likely caused by the fitting in high-dimensional space and the automatic determination of the number of components, which does not yield perfect results. However, compared to the visual representation of \uapca, \approach remains significantly closer to describing the underlying distribution, which explains why the sliced Wasserstein distance in \uapca is twice as high compared to \approach.

The Fashion-MNIST dataset contains 10 classes. \cref{fig:fashion_mnist_eval} shows the projections for both \uapca and \approach.
The detailed views to the right focus on four exemplary classes, allowing for a close examination of how the approaches represent the class structures.
While the ellipses based on \uapca describe the general distribution of the unimodal classes, the \approach provides a more faithful representation of the data, revealing deviations from normal distributions, such as skewed or elongated distributions.

\subsection{Evaluation of Analytic Student Grades Dataset}
\label{ssec:eval_students}
As an alternative application example, we consider the student grades dataset, which has been used in several uncertainty-aware dimensionality reduction approaches~\cite{denoeux2004fuzzy, Goertler2020UncertaintyAwarePCA, hagele2022uncertainty}. This dataset consists of four grades (M1, M2, P1, and P2) for six students, where the grades are associated with different types of uncertainty. For example, a grade could be given as ``fairly good'' or as a range $[10, 12]$. Each of the grades is modeled as either a trapezoidal distribution, a normal distribution, a uniform distribution, or is not associated with uncertainty, and no correlation is assumed among the different dimensions. We follow the definition as presented by H\"agele et al.~\cite{hagele2022uncertainty}. 

For modeling the distributions using GMMs, we first sample the dataset using 100,000 samples per student. This provides high-density sampling, enabling a very accurate GMM estimation. The GMM distributions are shown in~\cref{fig:student_eval}c. While the GMM matches the shape of the uniform and trapezoidal distributions better than a normal distribution, some deviations are visible.

\cref{fig:student_eval}c--e illustrate the projections with the different approaches that were also discussed previously. The projected distributions do not show ellipsoidal shapes, as can be seen in~\cref{fig:student_eval}c. A comparison with \uapca shows that the normal distributions fail to properly describe the piecewise linear functions of the student dataset. Projecting the GMMs (see~\cref{fig:student_eval}d) provides a visually better representation, which is also confirmed by the KL divergence and the sliced Wasserstein distance. However, the contour lines exhibit artifacts that can be attributed to the limited number of Gaussians used. 

\begin{figure}[!b]
    \centering
    \includegraphics[width=\linewidth]{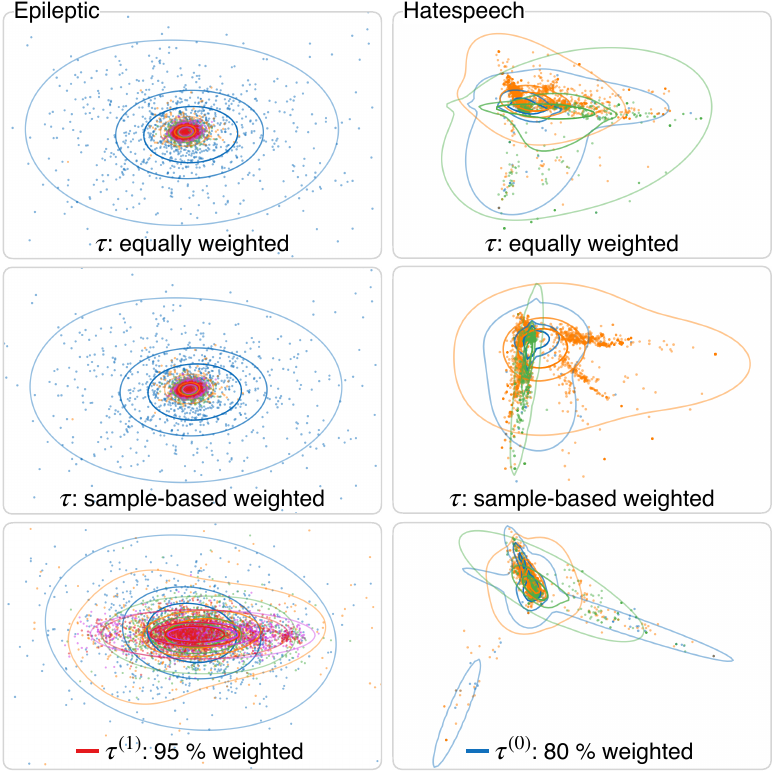}
    \caption{Projections of the epileptic (left) and hatespeech (right) datasets under different class weighting schemes: equal (top), sample-based (middle), and the class where the weights are shifted (bottom). One class receives most of the weight in the projection to reveal (possibly) hidden structures.}
    \label{fig:eval_weights}
\end{figure}

\subsection{User-Defined Weights}
\label{ssec:eval_weights}
In practice, classes in a dataset may differ significantly in their variance, sample size, or separation within the feature space. Applying equal or sample-based weighting in such cases can cause dominant classes to overshadow others, making subtle or compact structure harder to detect in the projected space. To overcome this problem, the weighting scheme can be adjusted to emphasize underrepresented classes, allowing their characteristics to emerge more~clearly.

To support the interactive exploration of user-defined weightings, we implemented a system that allows class weights to be adjusted. This system visualizes the resulting projection and contour plots in real time. A set of sliders provides individual control over each class weight, while ensuring that the total sum of weights remains normalized. The projection matrix is updated live upon interaction, allowing users to immediately see the influence of weight changes on the structure of the projected distributions. We use this system to analyze two datasets in detail: epileptic and hatespeech. These examples highlight two typical use cases for interactive reweighting.

In the epileptic dataset, weighting the classes differently compared to the default projection (see \cref{fig:eval_weights} (left)) reveals additional structure. Here, in both the equally weighted as well as the sample-based weighted projections, only the blue class (\textcolor{owndarkblue}{\textbf{---}}) is visible. In the user-specific weighting, the red class (\textcolor{ownred}{\textbf{---}}) was assigned a weight of 95\%, while the remaining weights are distributed evenly among the other classes. 
This modification enhances visual separability and, in this case, relaxes the former dense inner structure.

A second use case is the emphasis of a particular class that is of higher analytical interest, which may carry a higher semantic meaning. In the hatespeech dataset, the \textsc{`hatespeech'} class \mbox{(\textcolor{owndarkblue}{\textbf{---}})} may be of special importance due to its legal implications.
In \cref{fig:eval_weights} (right), a sample-based weighting reflects the natural label \mbox{distribution}. \mbox{Moving} the \textsc{`hatespeech'} class weight to 80\%, with the remaining 20\% split evenly between the other classes, shifts the projection focus and makes the structure of the emphasized class more prominent, potentially aiding targeted analysis.

User-defined weights thus provide a flexible and powerful mechanism for tailoring projections to analytical goals. The interactive visualization tool supports this by enabling exploration of weighting schemes and immediate visual feedback, helping users discover structures that would otherwise remain hidden.

\section{Discussion and Conclusion}
We have presented a method for projecting arbitrary probability distributions using a generalization of \uapca by modeling the uncertainty using GMMs.
The incorporation of weights, informed by dataset specifics, such as the sample size, or by domain knowledge of the user, allows for flexible control over the projection.
In both the quantitative and qualitative evaluation, we demonstrated the improved projection quality and usefulness for a wide range of real-world datasets. Furthermore, the application to arbitrary PDFs was presented for the student grades dataset.

Our evaluation used GMM fitting via expectation maximization, which is prone to converging to suboptimal solutions due to local maxima.
Therefore, the presented mixtures are not necessarily well fitted.
Using a \enquote{best of $n$} selection strategy, or a different fitting algorithm~\cite{kolouri2018slicedwsforgmm} would possibly improve our results.
While scalable in the number of distributions due to the aggregated moments, the selection and fitting of GMM components in high-dimensional space introduce additional complexity if they are not given in the first place.
In the future, we plan to investigate direct integrations based on \cref{eq:ansatz} to further improve the projection of distributions.

\section*{Appendix: PCA Projection of MVNs}
\label{appendix}

To validate the ansatz presented in \cref{eq:ansatz}, we apply it to a multivariate normally (MVN) distributed random variable $X\sim\mathcal{N}(\mu, \Sigma)$.
Substitution of the respective PDF
\begin{equation*}\label{eq:mvn_pdf}
    f_X(x)
= \frac{1}{(2\pi)^{D/2}\,|\Sigma|^{1/2}}\,
\exp\Biggl[-\frac{1}{2}(x-\mu)^T\Sigma^{-1}(x-\mu)\Biggr]
\end{equation*}
into the marginalization formula results in
\begin{align}
&f_Y(y) 
= \int_{\mathbb{R}^{D-d}} f_X\left(\begin{bmatrix} P & R \end{bmatrix} \begin{bmatrix}  y \\ z \end{bmatrix}\right)\,\diff z \notag \\ 
&= \frac{1}{(2\pi)^{D/2}\,|\Sigma|^{1/2}}
\int_{\mathbb{R}^{D-d}}\exp \Bigg[ -\frac{1}{2}
\bigg(
\left(
\begin{bmatrix} P & R \end{bmatrix} \begin{bmatrix} y \\ z \end{bmatrix} - \mu \right)^T \notag \\
&\quad \cdot\Sigma^{-1} \left(\begin{bmatrix} P & R \end{bmatrix} \begin{bmatrix} y \\ z \end{bmatrix} - \mu
\right)
\bigg)
\Bigg] \diff z\,. \tag{A.1}\label{eq:mvn_int} 
\end{align}

\noindent We divide the computation into the exponent and the normalization factor of the PDF. 

\paragraph{Exponent}
The main step used to marginalize the multivariate normal distribution is the usage of the Schur complement (SC) as shown in the lecture slides by Jordan~\cite[Sec. 13.3]{jordan_multivariate_2009}.
Since $U:=\begin{bmatrix} P & R \end{bmatrix}$ is orthogonal, the exponent in \eqref{eq:mvn_int} can be rewritten as
\begin{subequations}
\begin{align}
& \quad\;\exp \Biggl[ -\frac{1}{2}
\Biggl(
\left(
U \begin{bmatrix} y \\ z \end{bmatrix} - \mu
\right)^T
\Sigma^{-1}
\left(
U \begin{bmatrix} y \\ z \end{bmatrix} - \mu
\right)
\Biggr)
\Biggr] \notag \\
&=
\exp \Biggl[ -\frac{1}{2}
\Biggl(
\left(
U \begin{bmatrix} y \\ z \end{bmatrix} - \mu
\right)^T
U\,U^T \Sigma^{-1} U\, U^T
\left(
U \begin{bmatrix} y \\ z \end{bmatrix} - \mu
\right)
\Biggr)
\Biggr] \notag \\
&= \exp \,\Biggl[ -\frac{1}{2}
\Biggl(
\begin{bmatrix} y-P^T \mu \\ z - R^T \mu \end{bmatrix}^T
U^T \Sigma^{-1} U
\begin{bmatrix} y-P^T \mu \\ z - R^T \mu \end{bmatrix}
\Biggr) \Biggr]\,. \tag{A.2}\label{eq:exp_2}
\end{align}
\end{subequations}

\noindent Using $I=UU^T=U^TU$, we reformulate and define the middle factor in \cref{eq:exp_2}, $U^T \Sigma^{-1} U = (U^T \Sigma U )^{-1}=:(\Sigma')^{-1}$. Writing
\begin{equation*}
	\Sigma' = U^T\Sigma U = \begin{bmatrix} A & B\\ B^T & C \end{bmatrix}, 
	\quad
	A=P^T\Sigma P, \; B=P^T\Sigma R, \; C=R^T\Sigma R\,,
\end{equation*} the Schur complement gives
\begin{equation*}
 (\Sigma')^{-1}= 
\begin{bmatrix}
    I & -A^{-1} B \\ 0 & I
\end{bmatrix} 
\begin{bmatrix}
    A^{-1} & 0 \\ 0 & S^{-1}
\end{bmatrix} 
\begin{bmatrix}
    I & 0 \\ -B^T A^{-1} & I
\end{bmatrix}\,,
\end{equation*}
with $S = C-B^T A^{-1} B$. Thus, the exponent \eqref{eq:exp_2} can be written as
\begin{equation*}
    \exp\Biggl[-\frac{1}{2} \Big( (y-P^T \mu)^T A^{-1} (y-P^T \mu) + \tilde{z}^T S^{-1} \tilde{z} \Big)\Biggr] \, ,
\end{equation*}
where $\tilde{z}$ is defined as
$$
\tilde{z}  = z - R^T\mu - (R^T\Sigma P)\,(P^T\Sigma P)^{-1}\,(y - P^T\mu) \,.
$$
This isolates the $y$- and $z$-blocks, allowing the integrals to be split and the exponent to be integrated term by term.

\paragraph{Normalization}
Since $U$ is orthogonal, $\abs{\Sigma} = \abs{\Sigma'}$. Using the SC, the determinant factorizes to $\abs{\Sigma'} = \abs{A} \cdot \abs{S}$, leading to the split
$$
\frac{1}{(2\pi)^{D/2}\,|\Sigma|^{1/2}}
=
\frac{1}{(2\pi)^{d/2}\,\abs{A}^{1/2}}
\frac{1}{(2\pi)^{(D-d)/2}\,\abs{S}^{1/2}}
 \,.
$$

\noindent 
Combining the factorization of the exponent and the normalization, the integrand splits into a $y$-dependent Gaussian and a $z$-dependent Gaussian. 
Note that although $\tilde z$ depends on $y$, the change of variables is a translation in $\mathbb{R}^{D-d}$ with Jacobian $1$. Hence
\begin{equation*}
\int_{\mathbb R^{D-d}} 
\exp\!\Bigl(-\tfrac12 \tilde z^{\top} S^{-1}\tilde z\Bigr)\,\diff z
= (2\pi)^{\frac{D-d}{2}} \abs{S}^{1/2}.
\end{equation*}

\noindent
This cancels with the normalization factor 
$(2\pi)^{-(D-d)/2}\,|S|^{-1/2}$, 
so the $z$-dependent Gaussian integrates to $1$, leaving
\begin{align*}
\label{eq:mvn_result}
&f_Y(y)=\\
&\frac{1}{(2\pi)^{d/2}\,|P^T\Sigma P|^{1/2}}\,
\exp\Biggl[-\frac{1}{2}(y-P^T\mu)^T(P^T\Sigma P)^{-1}(y-P^T\mu)\Biggr]\,,
\end{align*}
or equivalently,
\begin{equation*}\tag{A.3}
Y \sim \mathcal{N}\bigl(P^T\mu,\,P^T\Sigma P\bigr) \,.
\end{equation*}
This matches the projection used in \uapca, showing our ansatz reduces to \uapca in the special case of multivariate normals.

\acknowledgments{\noindent A preliminary version of this work was done as part of a thesis by O.~Tastekin \cite{Tastekin2025}. We acknowledge funding by
the Deutsche Forschungsgemeinschaft (DFG, German Research Foundation) -- Project-ID 251654672 -- TRR 161.
}

\bibliographystyle{abbrv-doi-hyperref}
\bibliography{bib}

\end{document}